\documentclass[journal,twoside,web]{ieeecolor}
\usepackage{tmi}

\usepackage{cite}

\usepackage{caption}
\usepackage[caption=false,font=normalsize,labelfont=sf,textfont=sf]{subfig}

\usepackage{amsmath,amssymb,amsfonts}
\usepackage{bbm}
\usepackage{booktabs}
\usepackage{multirow}

\usepackage{hyperref}
\usepackage{algorithm, algorithmic}

\usepackage{todonotes}

\usepackage{inputenc}
\usepackage{graphicx}
\usepackage{makecell}

\usepackage{color}

\newcommand{\blueC}[1]{\textcolor[rgb]{0.00,0.00,0.00}{#1}}
\newcommand{\blueCN}[1]{\textcolor[rgb]{0.00,0.00,0.00}{#1}}
\newcommand{\blueCNN}[1]{\textcolor[rgb]{0.00,0.00,0.00}{#1}}

\usepackage{acronym}
\acrodef{RV}{Right Ventricle}
\acrodef{LV}{Left Ventricle}
\acrodef{MML}{Multiple Manifold Learning}
\acrodef{PLS}{Partial Least Squares}

\usepackage[british]{babel}

\def\BibTeX{{\rm B\kern-.05em{\sc i\kern-.025em b}\kern-.08em
    T\kern-.1667em\lower.7ex\hbox{E}\kern-.125emX}}
\begin{document}



\title{{Visualizing definitional divergence in high-dimensional data} by manifold alignment: \\ \LARGE{Application to 3D right ventricular strain computations}}

\author{Maxime~Di~Folco,
        Gabriel~Bernardino,
        Patrick~Clarysse,
        and~Nicolas~Duchateau
\thanks{M. Di Folco, G. Bernardino, P. Clarysse and N. Duchateau are/were with Univ Lyon, Universit\'e Claude Bernard Lyon 1, INSA-Lyon,CNRS, Inserm, CREATIS UMR 5220, U1294, F-69621, Lyon, France. E-mail: \url{maxime.difolco@telecom-paris.fr}}
\thanks{M. Di Folco \blueCNN{was} also with the Institute of Machine Learning in Biomedical Imaging, Helmholtz Center Munich, Germany\blueCNN{, and is with the LTCI, Telecom Paris, Institut Polytechnique de Paris, France}.}
\thanks{G. Bernardino is also with the DTIC, Universitat Pompeu Fabra, Barcelona, Spain.}
\thanks{N. Duchateau is also with the Institut Universitaire de France (IUF).}
\thanks{ACCEPTED FOR PUBLICATION IN IEEE TRANSACTIONS ON MEDICAL IMAGING. DOI: 10.1109/TMI.2026.3698240. \copyright~2026 IEEE. Personal use is permitted. For all other uses, permission must be obtained from IEEE.}}

\maketitle

\begin{abstract}

\blueCNN{Medical imaging studies often rely on a single sample per subject, assuming it is representative of their physiological traits. However, variations in how input descriptors are defined or computed (e.g. due to a lack of consensus in the scientific field) may have a crucial impact on the analysis, and are hardly considered in practice.} In this paper, we propose \blueCNN{an original strategy based on} representation learning to estimate \blueCNN{a parametric map reflecting the impact of such definitional differences on a given physiological descriptor, previously extracted from medical images.} \blueCNN{We consider the different definitions or computations of such physiological descriptors} as different high-dimensional data, potentially of heterogeneous \blueCNN{types. We} specifically focus on myocardial deformation (strain), for which \blueCNN{there is limited agreement on} its definition. We first use manifold alignment to match the latent representations associated with \blueCNN{the different definitions of this descriptor}. Then, we formulate plausible distributions \blueCNN{in the latent space to represent definitional divergence across descriptors, from which we reconstruct a high-dimensional parametric map to visualize such definitional divergence}. 

Due to the lack of proper ground truth \blueCNN{for this specific clinical application}, we first demonstrate this methodology on toy experiments and then expand the evaluation on right ventricular strain data from subjects obtained from 3D echocardiographic image sequences, for which different types of strain are available at each point of the right ventricle endocardial surface mesh. Beyond this illustrative application, our methodology has the potential to be generalised to many other population analyses considering heterogeneous high-dimensional descriptors.

\end{abstract}

\begin{IEEEkeywords}
Representation learning, uncertainty, information fusion, myocardial strain, cardiac imaging, 3D echocardiography.
\end{IEEEkeywords}


\section{Introduction}\label{sec:introduction}

\IEEEPARstart{D}{espite} advanced and relevant processing pipelines, medical image analysis methods suffer many times from two strong limitations: (i) the difficulty of establishing standards for some types of computations, 
and (ii) the blind acceptance of the input data as it is. Both issues can strongly affect the subsequent computations and therefore the disease analysis. The latter means that a single image is often considered representative of a patient's condition, while uncertainties on acquisition and measurements are well known. 
\blueCNN{In this paper, we explicitly target the former issue}, by proposing an original pipeline based on manifold alignment to match latent representations associated to different \blueCNN{definitions of a given input descriptor, and exploit the latent space to quantify the \blueCNN{discrepancy} associated to such definitional differences}. 

\vspace{2mm}

\subsection{Uncertainty modelling on high-dimensional data}

Uncertainties can arise at four different stages of medical image analysis with machine learning methods: data collection, data labelling, model selection, and model-based inference \cite{Begoli:2019}. Uncertainty sources are generally distinguished between aleatoric (typically, randomness in the data) and epistemic (namely, systematic) \cite{Hullermeier:2021}. The latter is directly linked with the design of machine learning models, and therefore receives most of the attention in the literature \cite{Kendall:NeurIPS:2017,Seoni:2023}, which also distinguishes between probabilistic and deterministic methods.

In this paper, we consider 
\blueCNN{the different ways a given data descriptor is defined or computed} (myocardial strain \blueC{from a given mesh sequence previously obtained from motion tracking})\blueCNN{. This notion does not explicitly correspond to aleatoric uncertainties, but 
also reflects} 
different views of the same patient data\blueCNN{.} In our cardiac imaging application, the strain patterns correspond to high-dimensional descriptors available at each point of the endocardial surface\blueCN{, and the strain patterns obtained from different definitions correspond to different types of descriptors.}
In this context, \blueCN{multimodal} dimensionality reduction methods are relevant to estimate an intermediate latent space of lower dimensionality in which we can represent these data and the correspondence between the different descriptors \cite{Kontolati:2022}\blueCNN{, which we exploit to consider the definitional divergence between such descriptors.}

The use of such intermediate latent space was introduced in geomathematics to estimate possible permeability maps from water flow data, but without explicitly quantifying uncertainty \cite{Caers:2010}. Explicit \blueCNN{estimation of aleatoric uncertainty} was obtained from the variance of high-dimensional descriptors reconstructed from the latent space, for example to localize myocardial infarct from 3D deformation \cite{Duchateau:TMI:2016} or model the subsurface of the Earth \cite{Blanc:TMI:2015}. In the former, uncertainty was modeled globally in the latent space, while the latter adjusted the size of the confidence regions (with probabilistic estimation) locally. In both cases, a single type of high-dimensional input data and (non-deep) manifold learning methods were used, meaning that reconstruction was done a-posteriori. More recently, \cite{Judge:MICCAI:2022} exploited latent space correspondences between two types of descriptors (images and segmentations) to model uncertainty on such segmentations. The latent space was estimated from an auto-encoder, and latent space correspondences were obtained from the CLIP method \cite{Radford:ICML:2021} originally designed to match image and text embeddings. 

Here, we go further by relying on a manifold alignment scheme (\ac{MML} \cite{Valencia:CIARP:2011,Lee:PR:2016,Clough:PAMI:2019}) that not only generalizes the alignment scheme to more than two descriptors, but also provides sample-wise flexibility depending on the link between the input descriptors. 
Manifold alignment is relevant as some data descriptors may have non-comparable numerical values but encode similar information, which should lead to aligned latent spaces\blueC{, namely samples with close latent coordinates. Conversely, descriptors that do not carry similar information should lead to samples with more distant latent coordinates.} 
\blueCN{It may be seen as a softer version of point cloud registration \cite{Monji:ISPRS:2023} and with dimensionality reduction performed at the same time.}
\blueC{It pursues objectives comparable to contrastive learning \cite{LeKhac:IEEEaccess:2020}, but within a manifold learning perspective, namely providing a latent space into which distances have statistical meaning. This notably allows navigating around existing coordinates as we propose to estimate \blueCNN{definitional divergence}.}
\blueC{Note that our purpose is not to build a consensus across several complementary descriptors, as in many multimodal approaches, most falling under the umbrella of fusion methods \cite{Baltrusaitis:PAMI:2019}, recently empowered with the attention mechanisms from transformers \cite{Xu:PAMI:2023}.}
\blueC{Also, achieving a perfect consensus is not possible, since some of the descriptors may contain partially different information.}
We \blueC{therefore} hypothesize that \blueC{the principles of manifold alignment} will allow relevant and rather direct sample-wise \blueCNN{modelling of definitional divergence} in the latent space, conditioned by the degree of matching between the input descriptors.

\vspace{2mm}

\subsection{Application: 3D \ac{RV} strain quantification from echocardiography}

The illustrative application we target here concerns the \blueC{computation} of 3D deformation (strain) of the \ac{RV} \blueC{from existing mesh sequences}. 
\blueCNN{Such computations generally involve a local reference frame based on}
three orthogonal directions derived from the geometry of the ventricle: radial (from endocardium to epicardium), circumferential (along the circumference), and longitudinal (from apex to base) \cite{Dhooge:EJE:2000}. This has advantages over using Cartesian coordinates, as this anatomical frame can be related to the directions of the myocardial fibers.
However, compared to the \ac{LV}, the \ac{RV} is particularly challenging because of its asymmetric shape \cite{Sanz:JACC:2019}: there is no consensus on the definition of these local directions for the \ac{RV} \cite{Duchateau:JASE:2021}, and 3D \ac{RV} strain quantification remains sensitive to differences in their definition and resulting computations. As a result, for echocardiography, clinical studies focus on 2D \ac{RV} strain quantified globally or regionally \cite{Badano:EHJCI:2018}, which leads to limited disease characterization. In 3D, coordinates-independent solutions consist in considering area strain (the relative area change of elements of the \ac{RV} surface) \cite{Moceri:EHJCI:2018}, or principal strain obtained by eigendecomposition of the strain tensor \cite{Satriano:JASE:2019}. Coordinates-dependent methods can consider global axes aligned to the main \ac{RV} dimensions \cite{Tokodi:FCV:2021}, or directions that mimic the definition given above \cite{Doste:IJNMBE:2019,Moceri:EHJCI:2021}, but even subtle differences can impact \ac{RV} local strain \cite{DiFolco:FIMH:2023}.

\vspace{2mm}

\subsection{Contributions}

We propose an original method to 
\blueCNN{consider definitional divergence (potentially involving heterogeneous data types) related to high-dimensional descriptors extracted from medical images.}
Manifold alignment matches the latent representations associated to the different high-dimensional input descriptors. \blueCNN{Definitional divergence is} modelled locally in the low-dimensional latent space, and \blueCNN{visualized after reconstruction through a high-dimensional parametric map.} We demonstrate \blueCNN{this approach} on experiments with toy and real datasets, using the data from a popular computer vision public dataset, and from a private dataset corresponding to a real cardiac imaging application where \blueCNN{the definition of standards for the input descriptors is} still a topic of debate \cite{Duchateau:JASE:2021,DiFolco:FIMH:2023}.

\begin{figure*}[t]
	\centering
    \includegraphics[width=0.95\textwidth]{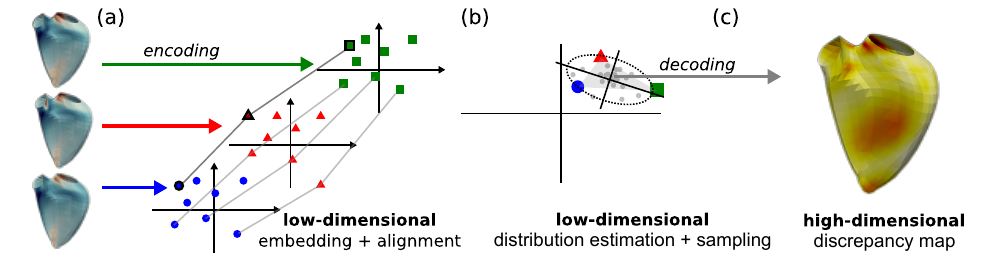}
	\caption{Overview of the pipeline proposed in this paper. (a) High-dimensional descriptors (here, \blueCNN{3D patterns corresponding to different definitions of \ac{RV} strain}) are encoded into low-dimensional latent spaces that are locally aligned depending on neighborhood relationships between samples (Sec.\ref{sec:ManifoldAlignment}). (b) Remaining differences in the latent space after alignment are exploited to \blueCNN{estimate a statistical distribution that models definitional divergence} (Sec.\ref{sec:UncertaintyModelling}). (c) Reconstructing 
    high-dimensional samples from this distribution \blueCNN{to visualize such definitional divergence} (Sec.\ref{sec:UncertaintyModelling}).} 
	\label{fig:Overview}
\end{figure*}

\section{Methodology}

Our methodology consists of three consecutive stages, as summarized in Fig.\ref{fig:Overview}:
\begin{itemize}
    \item[$\bullet$] The joint low-dimensional embedding of high-dimensional descriptors by manifold alignment (Fig.\ref{fig:Overview}a and Sec.\ref{sec:ManifoldAlignment}),
    \item[$\bullet$] The sampling of new low-dimensional points by estimating a distribution mimicking \blueCNN{definitional divergence} in the latent space for each subject (Fig.\ref{fig:Overview}b and Sec.\ref{sec:UncertaintyModelling}),
    \item[$\bullet$] \blueCNN{Reconstructing a high-dimensional parametric map} from the sampled low-dimensional points (Fig.\ref{fig:Overview}c and Sec.\ref{sec:UncertaintyModelling})\blueCNN{, referred to as the \emph{discrepancy map}}.
\end{itemize}
The  description of this generic approach is then complemented by insights on the 3D \ac{RV} strain data analyzed in this paper (Sec.\ref{sec:strain}). Our code is public\footnote{\url{https://github.com/maxdifolco/definitional_divergence}}, and includes a demo corresponding to the experiments on the toy data described in Sec.\ref{sec:coil-100}.

\subsection{\blueCNN{Quantification of definitional divergence}}
\label{sec:uncertainty}

\subsubsection{Manifold alignment}
\label{sec:ManifoldAlignment}

Given a population of $K$ subjects, we denote:
\begin{itemize}
    \item $\mathbf{X} = [\mathbf{x}_1, ... , \mathbf{x}_K]^T \in \mathbb{R}^{K \times D}$ the input data corresponding to one high-dimensional descriptor ($D$ being its dimensionality)\blueCN{: for example, the images associated to one channel of a set of RGB images as in Sec.\ref{sec:coil-100}, or the RV strain patterns associated to a given process to compute strain as in Sec.\ref{sec:LongStrain},}
    \item $\mathbf{Z} = [\mathbf{z}_1, ... , \mathbf{z}_K]^T \in \mathbb{R}^{K \times d}$ \blueCN{the low-dimensional latent coordinates corresponding to these data,} estimated by dimensionality reduction methods, with $d \ll D$.
\end{itemize} 

Given $M$ different descriptors of the same dimensionality $D$, we denote $\mathbf{x}_i^m$ and $\mathbf{z}_i^m$ the input data and low-dimensional coordinates associated to the $m$-th descriptor for the $i$-th subject, with $m \in [1,M]$.

\blueC{We perform manifold alignment through the }\ac{MML} algorithm \footnote{\scriptsize \blueCNN{\url{https://gitlab.com/maxDif/mml_pressure-volume-demo}}} \cite{Valencia:CIARP:2011,Lee:PR:2016,Clough:PAMI:2019}. \blueC{It consists in simultaneously estimating a low-dimensional latent space for each descriptor (first term in Eq.\ref{eq:MML_large}) while controlling, with a hyperparameter $\mu$, the alignment of the latent coordinates (second term in Eq.\ref{eq:MML_large}).} This amounts at minimizing:
\begin{equation}
    E(\mathbf{Z}) = \sum_{m=1}^M \sum_{i,j=1}^K \| \mathbf{z}_i^m - \mathbf{z}_j^m \|^2 W_{ij}^m + \mu \sum_{\substack{m,n=1 \\ m \neq n}}^M \| \mathbf{z}_i^m - \mathbf{z}_j^n \|^2 \blueCN{V_{ij}^{mn}},
    \label{eq:MML_large}
\end{equation}
where $\mathbf{W}^m = [ W_{ij}^m ] \in \mathbb{R}^{K \times K}$ stands for the affinity matrix of the $m$-th descriptor, and 
$\blueCN{\mathbf{V}^{mn} = [V_{ij}^{mn}]}  \in \mathbb{R}^{K \times K}$ 
encodes the correspondences between the $m$-th and $n$-th descriptors, weighted by a factor $\mu > 0$. \blueC{The elements of both matrices have values between 0 and 1.} Note that this formulation requires $\mathbf{z}_i^m$ and $\mathbf{z}_j^n$ to have the same dimensionality.

\blueC{The first term is similar to the low-dimensional embedding performed by Laplacian eigenmaps on each descriptor \cite{Belkin:2003}: two high-dimensional samples $\mathbf{x}_i^m$ and $\mathbf{x}_j^m$ that are close will lead to an affinity $W_{ij}^{m} \approx 1$, therefore forcing the low-dimensional coordinates $\mathbf{z}_i^m$ and $\mathbf{z}_j^m$ to be close. Conversely, two high-dimensional samples that substantially differ will lead to an affinity $W_{ij}^{m} \approx 0$, which will have no influence on bringing $\mathbf{z}_i^m$ and $\mathbf{z}_j^m$ close.}

\blueC{The second term operates in a comparable manner but considering cross-descriptor correspondences through 
$\blueCN{V_{ij}^{mn}}$, 
which constrains the distance between the inter-descriptor latent coordinates $\mathbf{z}_i^m$ and $\mathbf{z}_j^n$, namely the manifold alignment.}

In practice, $\mathbf{W}^m$ is defined through a Gaussian kernel, whose width $\sigma$ is set as the
average distance of each point to its $k_\sigma$-th neighbor. In addition to $\mu$ and $k_\sigma$, computations involve one additional hyperparameter, $k_M$, which stands for the amount of nearest neighbors to sparsify the extra-diagonal matrix 
$\blueCN{\mathbf{V}}$.

Computational details about the matrices involved in Eq.\ref{eq:MML_large} and its solution are given in Appendix \blueC{A}.

\vspace{2mm}

\subsubsection{Modelling definitional divergence}
\label{sec:UncertaintyModelling}

We \blueCNN{directly} exploit the low-dimensional latent spaces after manifold alignment\blueCNN{.} Concretely, \blueCNN{definitional divergence} is modelled sample-wise from the set of $M$ low-dimensional points $\mathcal{Z}^i = \{ \mathbf{z}^1_i, ... , \mathbf{z}^M_i \}$ associated to the $i$-th subject. If the neighbors of this subject are preserved across the different descriptors, then its low-dimensional coordinates $\mathcal{Z}^i$ should be close\blueCNN{. Conversly,} if the descriptors differ, \blueCNN{such coordinates should be distant, proportionally} by how much the descriptors differ. 

\blueCNN{Our methodology involves} a three-stage process\blueCNN{.}
We first estimate a multivariate Gaussian defined by the set of points $\mathcal{Z}^i$, whose variance and principal axes are obtained in practice by PCA on this set \blueCNN{(see illustration in Appendix B)}.
Our experiments involve two or three descriptors, meaning that $| \mathcal{Z}^i | = 2$ or $3$, and we therefore consider one or two main directions to define such Gaussian distribution, respectively. 
Then, we sample $N$ new low-dimensional points according to this distribution ($N=100$ in all our experiments).
\blueC{This value respectively corresponds to two times the recommended number of samples for estimating Gaussian distributions governed by 2 principal axes with a confidence level of $95\%$ (which corresponds to the case $| \mathcal{Z}^i | = 3$, the recommendation being lower for 1 principal axis, namely $| \mathcal{Z}^i | = 2$) \cite{Psutka:PR:2019}.}
Finally, we estimate the high-dimensional data associated to each of these new low-dimensional samples. \blueCNN{A high-dimensional parametric map reflecting definitional divergence is} obtained by dimension-wise standard deviation over the $N$ reconstructed high-dimensional samples (in our case, at each point of the 3D \ac{RV} mesh). 

Any of the $M$ descriptors can be reconstructed, although we mostly make observations on a single descriptor that stands as reference. In our case, we reconstructed the high-dimensional data by regression given that \ac{MML} does not explicitly have a decoding step. Specifically, we used multiscale kernel ridge regression\footnote{\scriptsize \url{https://github.com/nicolasduchateau/multiscale-kernel-regression}} \cite{Duchateau:GSI:2013} which is robust to the non uniform density of samples in the latent space. 



\subsection{Application to 3D RV strain}
\label{sec:strain}

Our \blueCNN{applicative focus is myocardial deformation} (3D strain, at each point of a 3D mesh of the \ac{RV}), \blueCNN{and the} different definitions or different ways to compute it.

\vspace{2mm}

\subsubsection{Data and pre-processing}

We processed RV surface meshes of 100 control subjects obtained from semi-automatic endocardial segmentation by an expert clinician and tracking of 3D \blueCNN{transthoracic} echocardiographic sequences using commercial software (4D RV Function 2.0, TomTec Imaging Systems GmbH, Germany). 
\blueCNN{More recent versions of this software exist, but this version corresponds to the one used in previous studies led by our clinical collaborator in which the studied subjects served as control cases \cite{Moceri:EHJCI:2018,Moceri:ECHO:2021,Moceri:EHJCI:2021}.}
\blueC{The protocol of these studies complied with the declaration of Helsinki and was approved by the local research Ethics committee. All subjects consented and provided written informed consent.}
\blueC{The 3D image sequences were acquired from an apical four-chamber view focused on the RV, using a matrix-array \mbox{X5-1} transducer (Philips Medical Systems). Care was taken to maximize the frame rate 
\blueCNN{($26 \pm 11$ Hz on this dataset)} 
and to include the entire RV within the images. The tracking software tracked the RV endocardial surface along the cardiac cycle using 3D speckle-tracking \cite{Muraru:EHJCI:2016}, and allowed exporting the sequence of 3D meshes for post-processing.}

The \blueC{RV surface} meshes were \blueC{handled} as VTK files, and consisted of 822 points and 1587 triangular cells,
after cropping out the tricuspid and pulmonary valves. The commercial software uses a mesh model that provides point-to-point mesh correspondences across subjects, attached to the mesh data. We also realigned them across the studied population using generalized Procrustes analysis with a rigid transform.

\blueCNN{The RV end-diastolic volume, RV end-systolic volume, and RV ejection fraction were $64.3 \pm 25.3$ mL, $29.2 \pm 13.3$ mL, and $55.0 \pm 6.3$ \%. Additional clinical characteristics about these patients can be found in the corresponding studies from our clinical collaborator \cite{Moceri:EHJCI:2018,Moceri:ECHO:2021,Moceri:EHJCI:2021}.}

\vspace{2mm}

\subsubsection{Local anatomical directions}

\begin{figure}[t]
	\centering
	\subfloat[\label{fig:Fig1a}]{
		\resizebox{0.32\linewidth}{!}{\includegraphics{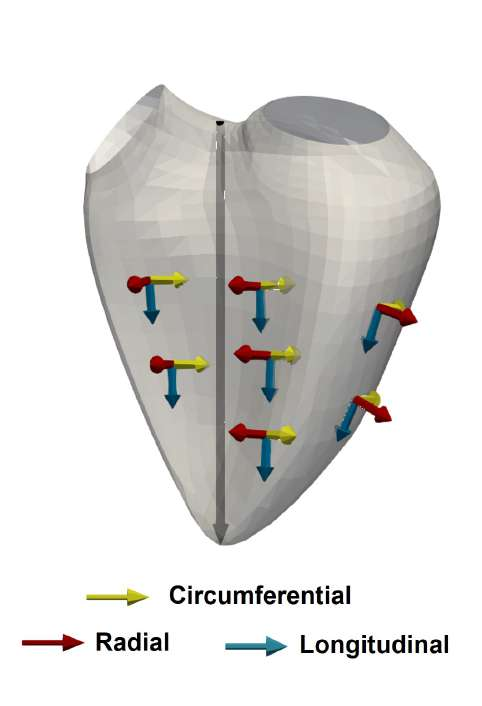}}}
	\subfloat[\label{fig:Fig1b}]{
		\resizebox{0.32\linewidth}{!}{\includegraphics{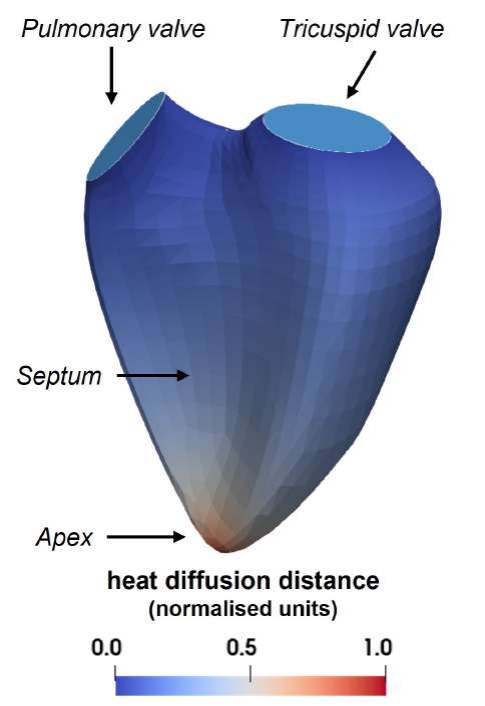}}}
	\subfloat[\label{fig:Fig1c}]{
		\resizebox{0.33\linewidth}{!}{\includegraphics{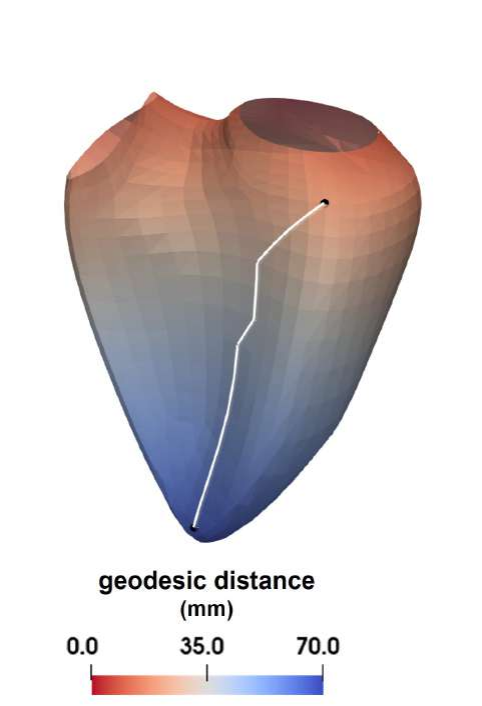}}}
	\caption{The three ways to compute circumferential and longitudinal directions we evaluated in this paper. (a) Long-axis computations, (b) Heat diffusion computations (the turquoise disks represent the cold point, while the apex stands as the hot point), (c) Geodesic distance computations (the white line corresponds to the geodesic joining the two purple dots).} 
	\label{fig:Fig1}
\end{figure}

As for the left ventricle, local directions can be defined at each cell of the \ac{RV} surface mesh, to be used afterwards to express strain along these directions. The radial direction is defined as the normal to the RV surface at each point. The longitudinal direction can be defined in three manners \cite{DiFolco:FIMH:2023} (Fig.\ref{fig:Fig1}):
\begin{itemize}
    \item[$\bullet$] \emph{Long-axis method:} \cite{Moceri:EHJCI:2018} The long-axis is defined as the segment joining the apex and the basal point equidistant from the valves centers. The circumferential direction is first estimated locally from the cross product between the radial direction and the long-axis. Then, the longitudinal direction is obtained from the cross product between the radial and circumferential directions.
    \item[$\bullet$] \emph{Heat diffusion method:} \cite{Doste:IJNMBE:2019} The apex and valves are defined as hot ($u=1$) and cold ($u=0$) points, respectively. The longitudinal direction is estimated as the gradient of the map $u$, defined at each point of the mesh by solving the partial differential equation $\nabla \cdot (\nabla u) = 0$. In our implementation, the map $u$ is estimated iteratively by (at each iteration) setting the value at each point as the weighted average of the values at neighboring points in the graph defined by the \ac{RV} mesh, updating all points, and then restoring the original values 1 and 0 to the apex and the valves.
    \item[$\bullet$] \emph{Geodesic distance method:} The longitudinal direction is defined as the gradient of the geodesic distance to the apex. We computed the exact geodesics in the discrete surface \cite{Mitchell:SJC:1987}. These do not necessarily follow the cell edges, and are therefore less prone to approximations compared to shortest path algorithms such as Dijkstra’s.
\end{itemize}
Finally, the circumferential direction is computed as the cross product between the radial and longitudinal directions.

\vspace{2mm}

\subsubsection{RV strain computations}

Once the local directions are computed, the local strain tensor is estimated as the Green Lagrangian strain:
\begin{equation}
    \mathbf{E} = \frac{1}{2} ( \mathbf{J}^T \cdot \mathbf{J} - \mathbf{I} ),
\end{equation}
where $\mathbf{J} = \nabla v + \mathbf{I}$, with $\nabla v$ the displacement gradient at a given point of the RV surface mesh, in Cartesian coordinates, and $\mathbf{I}$ is the identity matrix.

Then, longitudinal and circumferential directional strains are obtained by projecting the strain tensor along these two directions, as:
\begin{equation}
    \mathbf{E}_{\mathbf{h}} = \mathbf{h}^T \cdot \mathbf{E} \cdot \mathbf{h},
\end{equation}
where $\mathbf{h}$ is the unit vector defining the considered direction. As only endocardial \blueCNN{surface} meshes were available \blueCNN{for export from the commercial software used for segmentation and tracking}, radial strain was not \blueCNN{considered in this paper}.

In our database, all the strain patterns were available at each point of the \ac{RV} surface and at each instant of the cycle. Nonetheless, we focused the evaluation on end-systolic strain patterns, of higher magnitude. In all figures, results are displayed on end-diastolic meshes, which better render anatomical differences between subjects before deformation.

The algorithms to compute the local anatomical directions and 3D \ac{RV} strain of a 2D surface embedded in a 3D space were made publicly available\footnote{\scriptsize \url{https://github.com/gbernardino/rvmep}} following our publications dedicated to these aspects \cite{DiFolco:FIMH:2023,Bernardino:FIMH:2023}.

\vspace{2mm}

\subsubsection{Comparison schemes}
\label{sec:comparison}

We exploited the methodology described in Sec.\ref{sec:uncertainty} \blueCNN{on} \ac{RV} strain resulting from the following three configurations:
\begin{itemize}
    \item[$\bullet$] Longitudinal strain obtained from \underline{two} different computations of the local anatomical directions,
    \item[$\bullet$] Longitudinal strain obtained from \underline{three} different computations of the local anatomical directions,
    \item[$\bullet$] Longitudinal against circumferential strain.
\end{itemize}

As local directions and therefore strain computations may be influenced by the local \ac{RV} shape (as we observed earlier \cite{DiFolco:FIMH:2023}), we therefore specifically examined the link between the \blueCNN{reconstructed high-dimensional parametric map (referred to as the \emph{discrepancy map})}
across the whole RV surface and the shape onto which computation were made. To do so, we used the \ac{PLS} algorithm which serves to quickly relate two high-dimensional variables (in our case, the \blueCNN{discrepancy map} and the \ac{RV} shape). It provides a low-dimensional space whose main directions maximize the covariance between these two input descriptors, along which we can easily sample representative coordinates across the data distribution and reconstruct their corresponding high-dimensional data, in other words the \ac{RV} shapes that are most related to the main trends in \blueCNN{discrepancy maps}.



\begin{figure*}[ht!]
	\centering
	\includegraphics[width=0.8\textwidth]{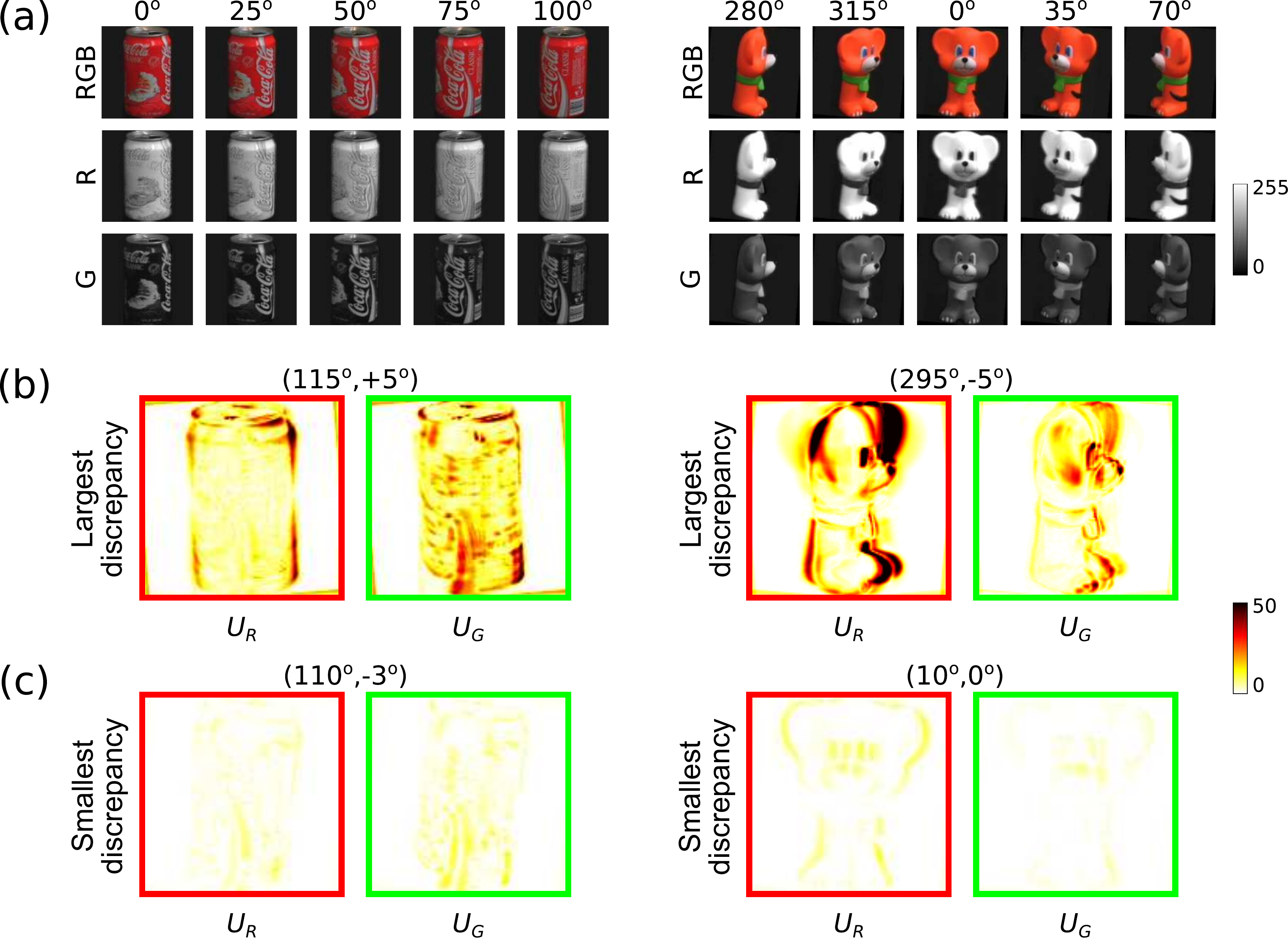}
	\caption{\blueCNN{Toy experiment involving} the red (R) and green (G) channels of images of two objects from the COIL-100 dataset \cite{Nene:1996}, of different view angles and additionally rotated between $-5^{\circ}$ and $+5^{\circ}$. (a) Samples of the two objects. The red channel dominates in both objects, being either almost uniform (soda can) or almost complementary of the green channel (toy bear). (b) Estimated \blueCNN{discrepancy map associated to definitional divergence, respectively for} the red and green channels, for the sample with the largest differences between the red and green latent spaces, after alignment with \ac{MML}. (c) Similar display for the sample with the smallest differences.}
	\label{fig:synthetic}
\end{figure*}

\section{Experiments}

\subsection{Implementation details}

For all experiments, $N$, the number of points sampled in the latent space to \blueCNN{model definitional divergence} is set to 100. For the experiment on the COIL-100 dataset (Sec \ref{sec:coil-100}), \ac{MML} was used with the following parameters: $\mu=1$, $k_\sigma = 5$, and $k_M = 5$. Concerning the experiment on the \ac{RV} strain (from Sec \ref{sec:noise} to Sec \ref{sec:long_circ}), MML was computed with $\mu=1$, $k_\sigma = 10$ neighbors and $k_M = 10$.
\blueCNN{These values are similar to those determined in our previous work \cite{DiFolco:MedIA:2022}, which used MML on 3D RV meshes from the same commercial software. 
This work included a sensitivity analysis of the influence of such hyperparameters on the learnt latent representation for synthetic data (left ventricular pressure and volume curves). In brief, for the RV meshes data, $k_\sigma$ was set to around $10\%$ of the available samples. At fixed $k_M$, the optimal $\mu$ was chosen as the one minimizing the energy defined in Eq.\ref{eq:MML_large}. $k_M$ was empirically determined by examining the latent representation.}

\blueCNN{Computations were fast. On a laptop
with Intel Core 4.6 GHz i5-1335U processor and 32 Gb memory, executing the whole pipeline (Matlab code) took less than 10 s for each of the two objects from the COIL-100 dataset, and for each experiment on RV strain patterns from the 100 control subjects.}

\subsection{Toy data: color channels in RGB images}
\label{sec:coil-100}
We first evaluated our approach on image samples from the COIL-100 dataset \cite{Nene:1996} which consists of RGB images of 100 objects, each object being observed under the 72 same viewpoints (a set corresponds to a 360-degree rotation of the object around itself). The three channels of an RGB image can be seen as three different but related ways to represent the observed object, which we may consider as different descriptors of this object. 
In this experiment, we used our method \blueCNN{to apprehend the definitional divergence} associated with representing a given object from two of these different channels. 
\blueCN{The \blueCNN{discrepancy map linked to} a given channel (e.g. red) corresponds to the error in reconstructing this channel from a latent representation that does not exactly correspond to this channel (e.g. latent coordinates between the red and the green latent coordinates.} 

We picked two datasets for which the two channels provide complementary information on a large enough portion of the object, such as the soda can (where the red channel is almost uniform, while the green channel is almost present on the brand name) and the toy bear (where the red and green channels are almost complementary except around the nose). We restricted the view angles to prevent redundancies in the images and circular distributions of latent coordinates (from $0^{\circ}$ to $90^{\circ}$ and $270^{\circ}$ to $355^{\circ}$ for the toy bear, from $0^{\circ}$ to $115^{\circ}$ for the soda can, each by steps of $5^{\circ}$). To benefit from more samples and wider distribution of samples in the latent space, we added rotated images of each existing samples (between $-5^{\circ}$ and $+5^{\circ}$ by steps of $1^{\circ}$). This led to 264 and 407 samples for the soda can and toy bear, respectively.

Figure \ref{fig:synthetic} displays the original RGB images and analyzed red and green channels at representative viewpoints, and the \blueCNN{discrepancy maps} estimated for each channel (denoted $U_{R}$ and $U_{G}$), obtained after aligning the red and green channel latent spaces. For the soda can, \blueCN{the two channels do not exactly carry the same information and therefore each channel \blueCNN{impacts} the reconstruction of the other channel. Indeed, the red channel is rather uniform within the object and therefore represents less the object rotation compared to the green channel. Regarding reconstruction, given the uniformity of the red channel,}
\blueCNN{the highest values in the discrepancy map} are mostly concentrated at the borders of the object. Conversely, the green channel mostly contains information near the brand name and the Santa drawing, and \blueCNN{the highest values in the discrepancy map} are concentrated near the borders of these locations. 
\blueCN{A comparable interpretation can be given for the toy bear. Both red and green channels carry information about the object rotation, as the orientation of the object within the images (and its pixel intensities) varies across the different samples. However, this concerns the whole object for the red channel, while this only concerns part of the head for the green channel (namely, less accurate information about the whole object rotation). Regarding reconstruction, as the red channel is rather uniform except around the neck, \blueCNN{the highest values in the discrepancy map} are mostly around the borders of the object. They} are also located around this zone for the green channel, and not around the neck despite some information on this channel, as most of the necklace (except its right extremity) remains unchanged when rotating the view angle.

\subsection{Toy data: \ac{RV} strain vs. noisy \ac{RV} strain}
\label{sec:noise}
\begin{figure*}[ht!]
	\centering
	\includegraphics[width= 0.75 \linewidth]{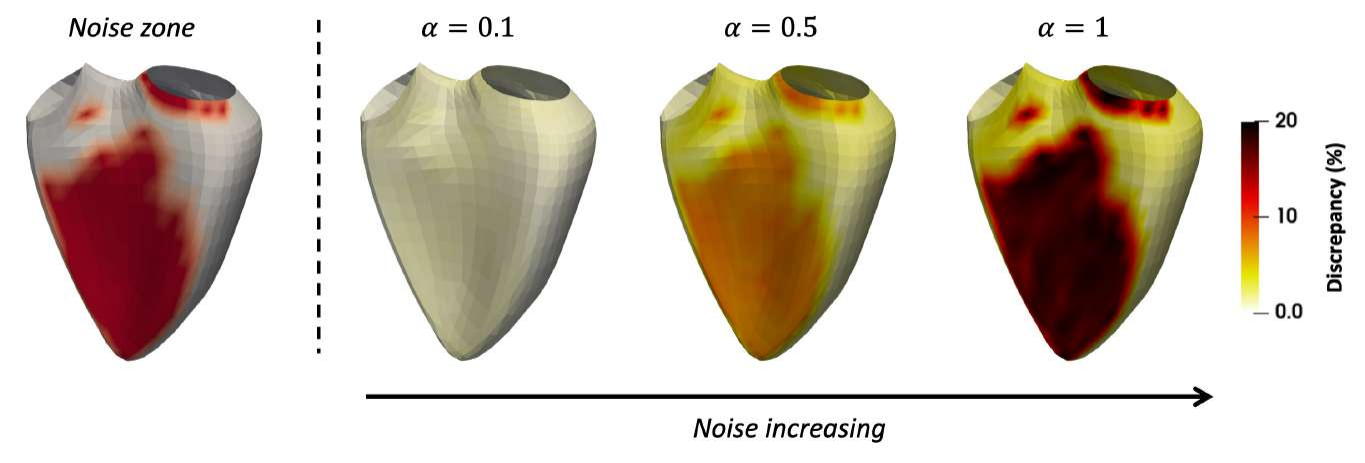}
	\caption{\blueCNN{Toy experiment involving} as high-dimensional observations the original RV strain ($\{ \mathbf{x}_i^0 \}$) and an artificially noisy version of it ($\{ \mathbf{x}_i^1 \}$). Noise was introduced on the high-dimensional RV strain in a given zone (see left side of the figure), \blueCN{and we tested} different noise intensity levels ($\alpha \in [0,1]$).}
	\label{fig:noisy_exp}
\end{figure*}

We also evaluated our method on another toy dataset, using the \ac{RV} strain from our medical application. The representation was estimated from the whole population of 100 control subjects. The two descriptors \blueCN{(namely, the high-dimensional observations $\{ \mathbf{x}_i^0 \}$ and $\{ \mathbf{x}_i^1 \}$)} were the original \ac{RV} strain (estimated using the \textit{long-axis} method) at each point of the mesh, and the same \ac{RV} strain but with artificially introduced noise in a specific region of the \ac{RV} (same region for all subjects). Gaussian noise ($\mathcal{N}(0,100)$) was added to the original strain values \blueCN{(namely, to the $\{ \mathbf{x}_i^1 \}$)}, controlled by a scaling factor $\alpha$. Figure \ref{fig:noisy_exp} summarizes our observations. The left part of the figure indicates the zone where noise was added, while the other miniatures depict the \blueCNN{discrepancy maps for} $\alpha = 0.1$, $0.5$, and $1$ ($\alpha = 1$ actually means substantial noise compared to the peak strain magnitude around $0.5$, see Figs.\ref{fig:sample_full} and \ref{fig:mean}). \blueCNN{The values of the discrepancy map increase} according to the value of $\alpha$, mainly in the zone where noise was added. The non-zero \blueCNN{values} outside of the noisy zone are both due to the sampling process in the latent space, which induces variations on the whole high-dimensional strain pattern and therefore across the whole \ac{RV}, and to the regression, which averages neighboring patterns. Despite these limitations, our method is able to capture \blueCNN{definitional divergence in a relevant manner} for all values of $\alpha$. 

\blueCNN{To complement these qualitative observations with quantitative measurements, w}e generalized this experiment \blueCNN{to} the 8 different zones into which clinicians commonly divide the \ac{RV} surface, defined in \cite{Haddad:Circ:2008}. We added noise with $\alpha=1$ in each of these zones, separately, and quantified the \blueCNN{average of the discrepancy map} per zone for the whole population (Tab.\ref{tab:noisy_exp}). The last row corresponds to the \blueCNN{average of the discrepancy map} obtained between the \textit{long-axis} and \textit{heat diffusion} computations, for comparison purposes. We observe that \blueCNN{the estimated values are} always more important in the noisy zone compared to the other ones. A \blueCNN{global} increase is also observed compared to the \emph{long-axis} vs. \emph{heat diffusion} values, mostly because the range of values with $\alpha = 1$ leads to larger strain values.

\begin{table}[h]
	\centering
	\resizebox{0.5\textwidth}{!}{\scalebox{1.3}{\begin{tabular}{l c c c c c c c c c }
\toprule
~ & ~ & \multicolumn{8}{c}{\blueCNN{Average of the discrepancy map} per zone} \\
~ & ~ & $\#$1 & $\#$2 & $\#$3 & $\#$4 & $\#$5 & $\#$6 & $\#$7 & $\#$8 \\ 
\midrule

\multirow{8}{*}{\rotatebox{90}{Noisy zone}} & 
$\#$1 & \textbf{17.56}  &  4.07  &   2.96  &    2.32  &  6.10  &  3.31   &  2.53   &  1.85  \\
~ & $\#$2 &   5.62  &  \textbf{16.95}   &  4.51   &   2.68 &  3.83  &   3.94  &    2.80  &    3.15 \\
~ & $\#$3 &   4.36  &  6.74   &  \textbf{17.96}   &   6.50 &  5.59  &   6.25  &   3.50  &    2.24 \\
~ & $\#$4 &   2.79  &  2.82   &  6.40   &   \textbf{18.04} &  5.27  &   4.92  &   7.03  &    3.45 \\
~ & $\#$5 &   8.35  &  3.40   &  3.91   &   3.86 &  \textbf{18.45}  &   6.68  &   5.92  &    2.92 \\
~ & $\#$6 &   2.60  &  2.61   &  3.60   &   3.05 &  5.51  &    \textbf{17.96}  &   5.04  &    3.96 \\
~ & $\#$7 &   2.10  &  1.97   &  2.67   &   4.70 &  5.84  &   6.22  &   \textbf{17.57}  &    6.01 \\
~ & $\#$8 &   2.13  &  3.37   &  2.53   &   3.25 &  3.71  &   5.88  &     7.03  &   \textbf{ 17.49} \\

\midrule
\multicolumn{2}{c}{Baseline} &  0.68 & 0.66 & 1.05 & 0.90 & 1.13 & 1.35 & 1.13 & 0.79 \\
\bottomrule
\end{tabular}}}
\vspace{2mm}

%
%
%
%
%
%
	\caption{\blueCNN{Toy experiment involving} noisy ($\alpha = 1$, in a given zone of the \ac{RV}) and original \ac{RV} strain patterns (\emph{long-axis} computation). Values correspond to the average \blueCNN{of the discrepancy map}, in $\%$ (strain units), across the population of 100 control subjects. The baseline corresponds to the \blueCNN{discrepancy map} estimated between the \emph{long-axis} and \emph{heat diffusion} computations. \\ {\scriptsize RV zones: $\#1$: outflow tract, $\#2$: anterior wall, $\#3$: lateral wall, $\#4$: inferior wall, $\#5$: infundibular septum, $\#6$: membranous septum, $\#7$: inlet septum, $\#8$: trabecular septum.}}
	\label{tab:noisy_exp}
\end{table}

\subsection{Longitudinal strain}
\label{sec:LongStrain}

\blueCNN{The} previous toy experiments demonstrated that our approach is able to identify a relevant zone \blueCNN{related to definitional divergence} and return higher \blueCNN{values} in case of higher differences between the input descriptors. In this section, we compare the \blueCNN{discrepancy maps} associated to the different computations of longitudinal strain described in Sec.\ref{sec:strain}: \textit{long-axis}, \textit{heat diffusion} and \textit{geodesic}. We chose the \textit{long-axis} as reference computation and \blueCNN{estimated the discrepancy map} on this descriptor when also considering one of the two others strain computations as second descriptor, or considering all three together.

\begin{figure*}[ht!]
    \centering
    \includegraphics[width=0.89\linewidth]{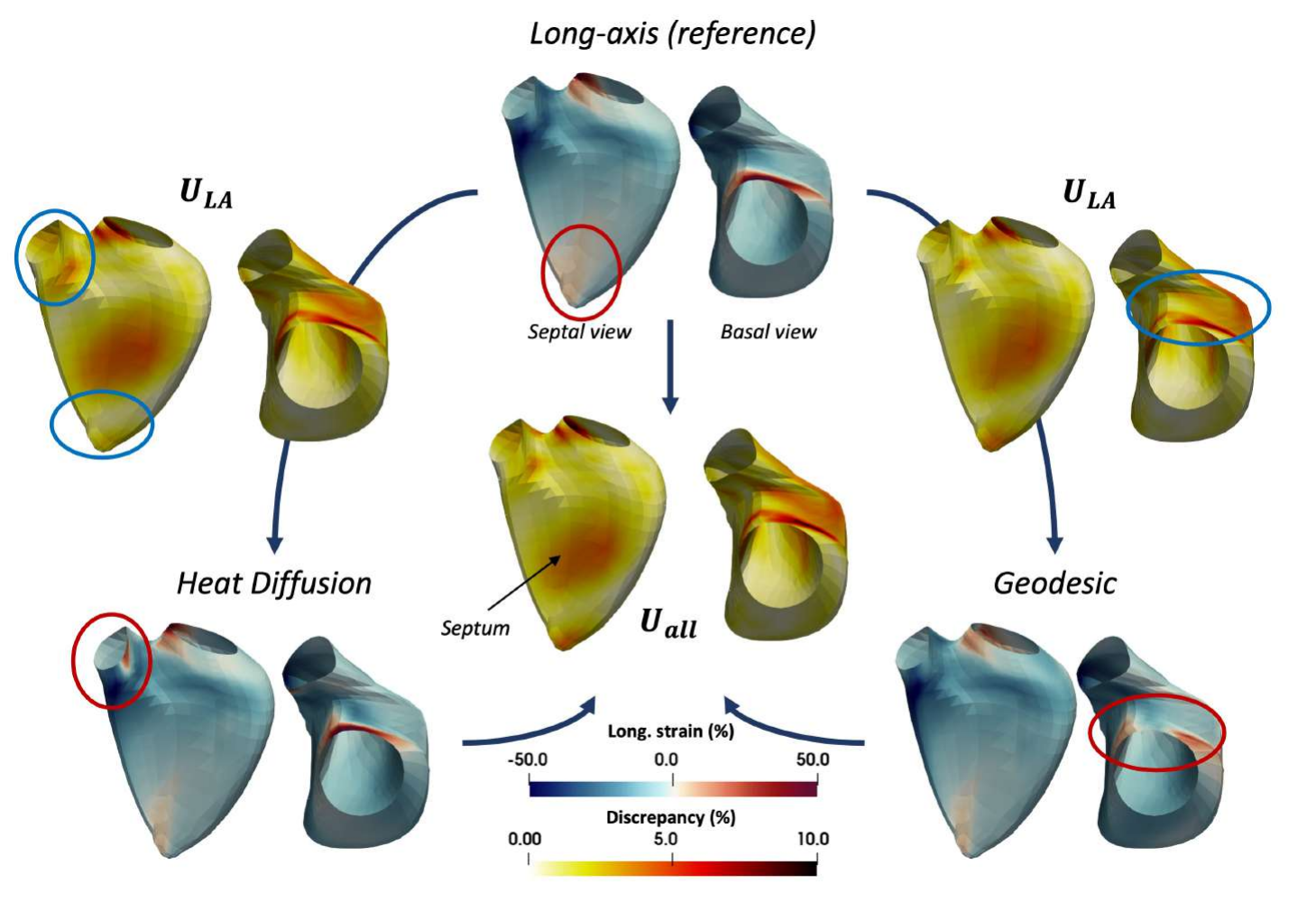}
    \caption{Strain patterns from a representative subject, and corresponding \blueCNN{discrepancy maps}. The Long-axis (LA) is used as reference for comparison between two descriptors (with \emph{heat diffusion} and \emph{geodesic} computations) and all three descriptors together (center of the figure). The red and blue circles highlight zones of major differences for \blueCNN{the} strain and \blueCNN{discrepancy maps}, respectively.}
    \label{fig:sample_full}
\end{figure*}

Figure \ref{fig:sample_full} displays the strain pattern estimated from the three different types of computation for a representative subject, and the corresponding \blueCNN{discrepancy maps}.


The strain and \blueCNN{discrepancy maps} are quite similar in general, except for the zones marked with red (strain) and blue (\blueCNN{discrepancy}) circles, respectively. Between the valves, the \emph{geodesic} strain is different from the \emph{long-axis} strain, which is correctly captured \blueCNN{in the discrepancy map}. Similarly, the \textit{heat-diffusion} strain pattern is slightly more expanded and pronounced in this zone, as also visible on the \blueCNN{discrepancy map} between these two descriptors. Comparable observations can be made for the other zones marked with a circle. Finally, \blueCNN{the discrepancy map} resulting from the analysis of the three strain descriptors together kind of merge the previous observations in a single map.


\begin{figure*}[ht!]
    \centering
    \includegraphics[width=\textwidth]{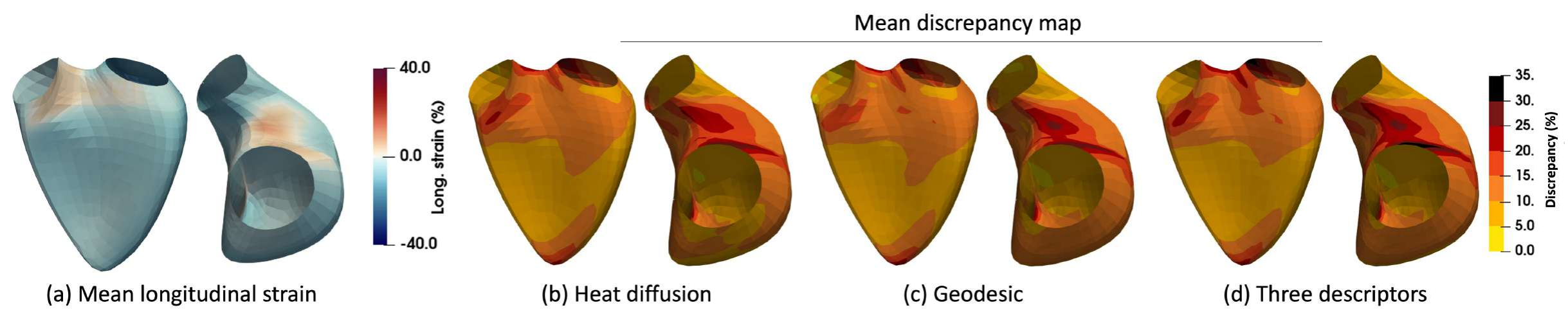}
    \caption{\blueCNN{Average discrepancy map} (b-d) across the population for the different strain computation schemes described in Sec.\ref{sec:comparison}. (a) Average longitudinal strain pattern of the \emph{long-axis} computation, used as reference in \blueCNN{this experiment}.}
    \label{fig:mean}
\end{figure*}

\begin{figure}[ht!]
	\centering
	\includegraphics[width=\columnwidth]{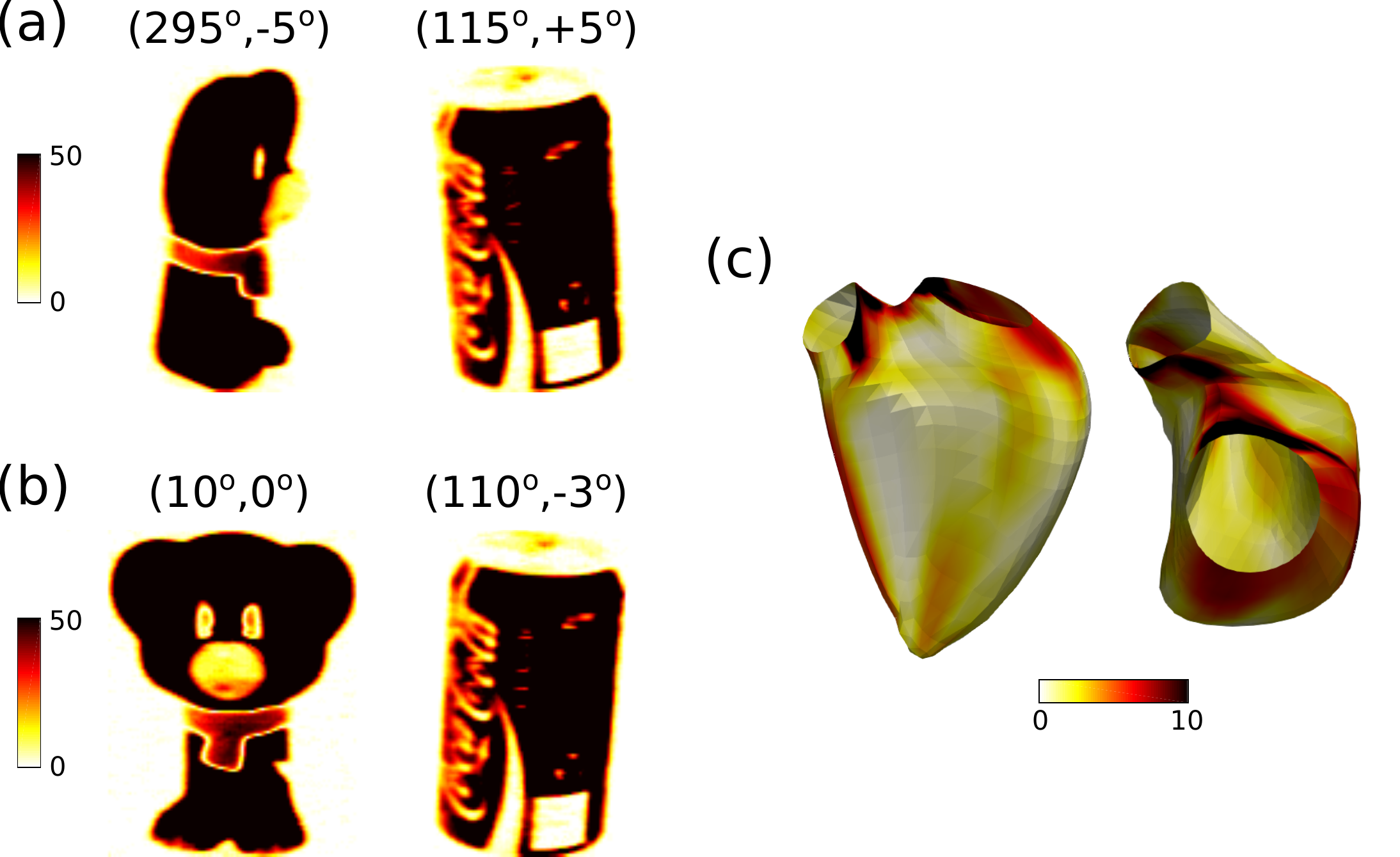}
	\caption{Standard deviation between the input descriptors. (a) and (b) Red and green channels, for the soda can and toy bear at the orientations shown in Fig.\ref{fig:synthetic}b and c. (c) All the three strain descriptors, for the subject depicted in Fig.\ref{fig:sample_full}.}
	\label{fig:std}
\end{figure}

We then computed the mean \blueCNN{discrepancy map} across the population for the three different types of strain computation (\emph{long-axis} either vs. \emph{heat diffusion} or \emph{geodesic}, and the three descriptors together), and displayed it on the average mesh for the population, and end-diastole (Fig.\ref{fig:mean}). We discretized the color code to enhance the visualization of \blueCNN{distinct} zones. We first observe that \blueCNN{the displayed maps} are quite similar across the three types of strain computation, with \blueCNN{the highest values} around the valves and near the apex. Local \blueCNN{high values} are observed under the pulmonary valve in Fig.\ref{fig:mean}b and between the valves in Fig.\ref{fig:mean}c). These also appear when the three descriptors are considered together. This experiment shows that our approach correctly \blueCNN{reflects definitional differences between} each type of computation when more than two input descriptors are used.

Of note, the \blueCNN{high values of the discrepancy map} for all of the experiments \blueCNN{are} concentrated in a zone that is difficult to track in 3D echocardiography (close to the apex and the valves), also meaning that \blueCNN{the maps we estimate} might be correlated to the local shape of the \ac{RV}. This is what we specifically evaluate in \blueCNN{Sec.\ref{sec:shape_uncertainty}}.


\subsection{Comparison with the standard deviation between the input descriptors}

One may wonder if the manifold alignment and exploitation of the latent space are really \blueCNN{necessary, compared} to a much simpler approach of directly computing the standard deviation between the input descriptors for each sample (if the different types of such descriptors allow it). This is illustrated in Fig.\ref{fig:std}, both for the toy data from Sec.\ref{sec:coil-100} and for the longitudinal strain data from Sec.\ref{sec:LongStrain}. 
\blueCN{On the toy data,} \blueCNN{the discrepancy map was} overestimated with this approach\blueCN{, both regarding magnitude and spatial distribution. First, \blueCNN{high values are} spread almost uniformly across the whole object,} as the red and green channels do not \blueCN{have comparable intensities} at the same locations (e.g. almost no green color over the can and bear body). \blueCN{However, both channels encode the object rotation (although with slight differences), and \blueCNN{the impact of} a given channel on the other one should therefore be lower, as previously observed in Fig.\ref{fig:synthetic}.}
Besides, \blueCN{in Fig.\ref{fig:std},} the \blueCNN{displayed values have comparable magnitude} for the two displayed orientations, while these were actually associated to the largest/smallest discrepancies in Fig.\ref{fig:synthetic}, due to different grades of manifold alignment around these samples. The strain data is more subtle to assess, but we also observe an overestimation of \blueCNN{the discrepancy map} both regarding magnitude (e.g. near the valves) and \blueCN{spatial distribution} (e.g. near the septal/lateral wall junction or near the apex).

This simple experiment confirms the relevance of manifold alignment: besides different numerical values, the input descriptors may encode similar information for some samples, which is insufficiently captured by directly computing the standard deviation between the input descriptors.

\subsection{\blueCNN{Link with the RV shape}}
\label{sec:shape_uncertainty}

\begin{figure}[ht]
    \centering
    \includegraphics[width=1\linewidth]{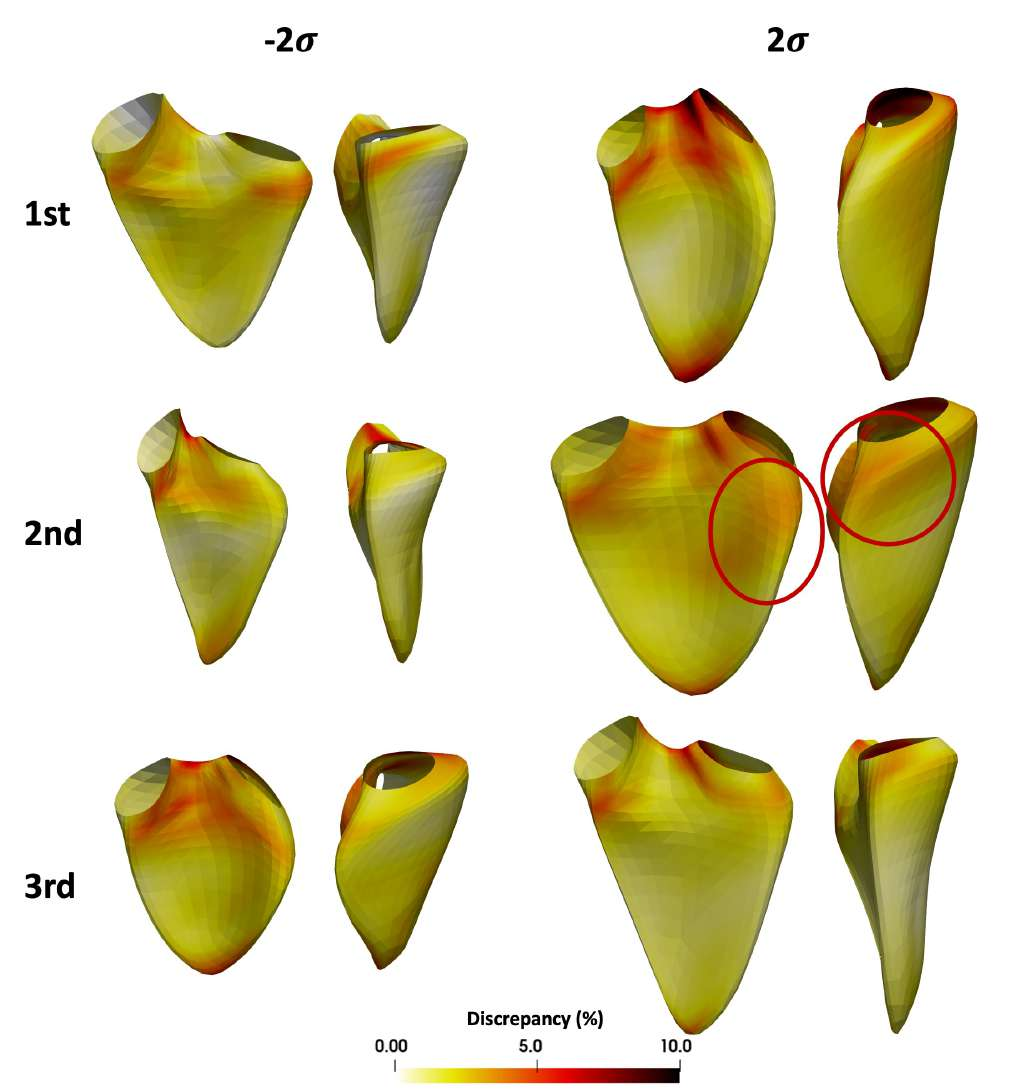}
    \caption{First three modes of variation of \blueCNN{discrepancy maps} and \ac{RV} shapes obtained from PLS. $\sigma$ stands for the standard deviation along the corresponding dimension.}
    \label{fig:vM_PLS}
\end{figure}

In this section, we examine the link between the \blueCNN{discrepancy maps} obtained using the three descriptors related to the different strain computations, and the local \ac{RV} shape, using the experimental setup based on \ac{PLS} and described in Sec.\ref{sec:comparison}. Shape was considered through the set of 3D coordinates at each point of the mesh. Figure \ref{fig:vM_PLS} shows the first three joint modes of variation of shape and \blueCNN{definitional divergence}, estimated by reconstructing shape and \blueCNN{discrepancy maps} at $-2\sigma$ and $+2\sigma$ along each dimension estimated by \ac{PLS}, with $\sigma$ the standard deviation of the corresponding dimension. 

The first mode of variations mostly encodes the relative position of the valves (tricuspid vs. pulmonary valves). High \blueCNN{values in the discrepancy map are} observed under the pulmonary valve when the tricuspid valve is more elongated. This mode of variation also encodes the length of the \ac{RV}, with \blueCNN{high values} near the apex when the \ac{RV} is long. 
The second mode of variation corresponds to global volume and septal curvature (observable in the costal view). An increase of \blueCNN{values} in the zones marked by the red circles is observed when the curvature increases. Both the second and third modes of variation show larger and more round shapes with \blueCNN{high values of the discrepancy map} under the valves. 

While variations in volumes and lengths are expected with a linear approach like \ac{PLS}, most shape variations align with regions challenging to track in 3D echocardiography: the apical septum and areas near the valves. Defining anatomical directions in these areas is difficult, resulting in distinct strain patterns between the different computations, leading to high \blueCNN{discrepancies}. 

\subsection{Longitudinal vs. Circumferential strain}
\label{sec:long_circ}
\begin{figure}[ht]
    \centering
    \includegraphics[width=1\linewidth]{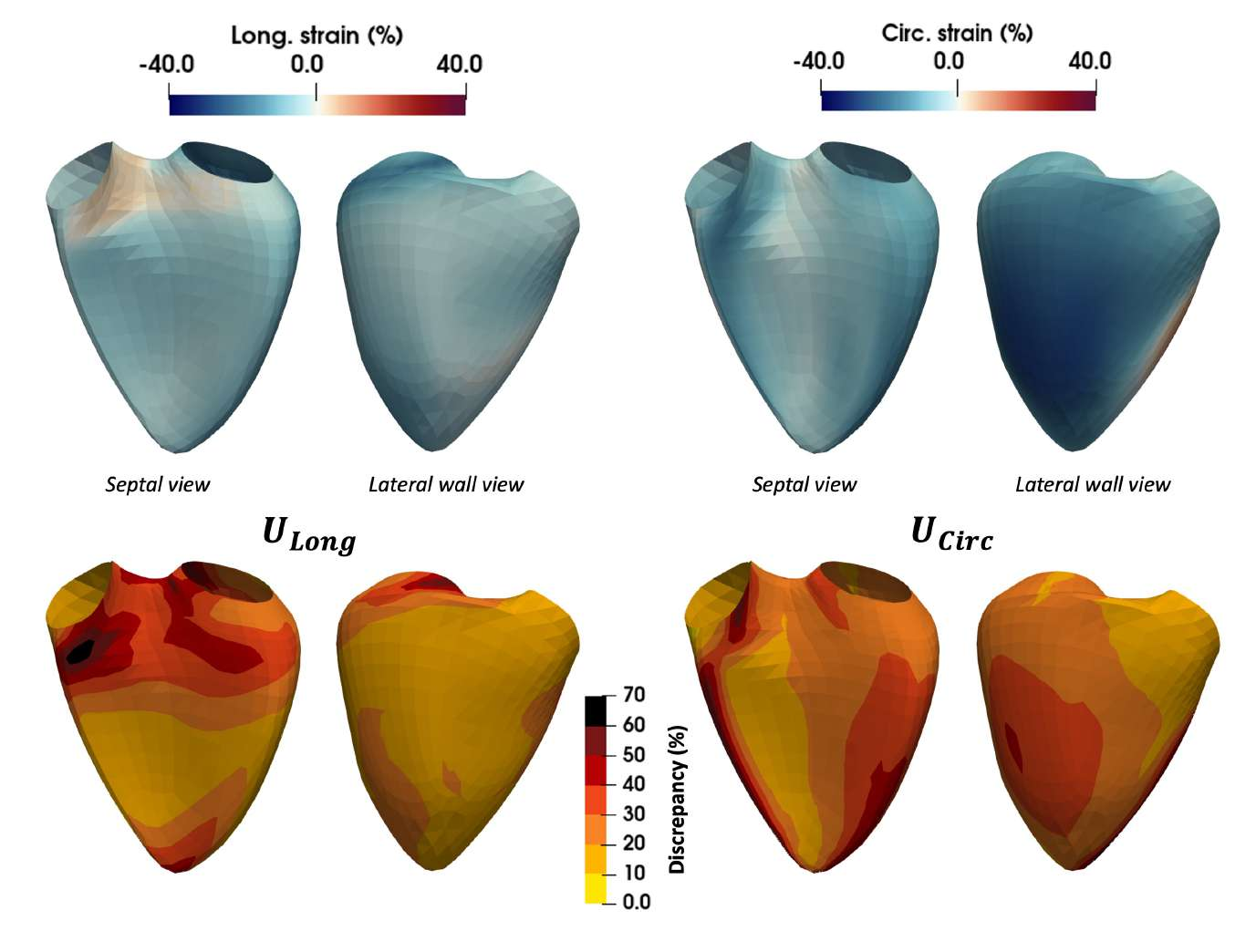}
    \caption{Average strain (top row) and \blueCNN{discrepancy maps} (bottom row) for the longitudinal (Long.) and the circumferential (Circ.) strain.}
    \label{fig:mean_long_circ}
\end{figure}

In this section, we \blueCNN{extend the evaluation to} longitudinal and circumferential strain using the \emph{long-axis} computations. In contrast to the previous experiments on the \ac{RV}, the two descriptors represent different information as visible from the mean strain patterns in Fig.\ref{fig:mean_long_circ}. Circumferential strain is higher on the free wall and near the border between the septum and the free wall, while longitudinal strain is higher around the valves and the apex\blueCNN{, leading to different observations}. The $U_{Long}$ pattern is similar to the one observed in Fig.\ref{fig:mean}, indicating that despite considering another descriptor, our method is able to capture the variations of the descriptor individually. Concerning the $U_{Circ}$ pattern, the zone of high \blueCNN{values} corresponds to the area with high strain, which is consistent with the observations made for longitudinal strain. 

\section{Discussion}

In this paper, we proposed an original methodology to \blueCNN{apprehend definitional differences for} high-dimensional descriptors extracted from imaging \blueCNN{data.} Our methodology exploits representation learning and in particular manifold alignment, which puts in correspondence the latent spaces associated to different input descriptors, these descriptors being seen as different versions of the input data. The \blueCNN{definitional divergence is considered from} the amount of (mis)alignment in the latent space, directly linked to the differences in information carried out by the different descriptors.

We demonstrated the relevance of our method on two toy experiments. First, on some samples from the COIL-100 dataset of RGB images, where \blueCNN{the estimated discrepancy map} corresponded to differences in the information encoded by some RGB channels. Then, on synthetic data generated from real \ac{RV} strain patterns with additional noise in a given region \blueCNN{(which enables quantitative assessment, and the noisy region can stand as ground truth location), the discrepancy map} actually reflected the noise level introduced in such region.
Finally, we evaluated the soundness of our method on real \ac{RV} strain patterns from 100 control subjects, where \blueCNN{definitional divergence corresponded} to different ways to compute local anatomical coordinates across the \ac{RV} and therefore myocardial strain. Our approach was able to consider more than two descriptors at once, and the estimated \blueCNN{discrepancy maps} were consistent with those estimated from pairs of descriptors taken independently, and could be interpreted in light of representative shape variations.

Analysing the mean \blueCNN{discrepancy map locally} confirms that \blueCNN{the highest values were found in} regions that are challenging to define anatomical directions, i.e. the apical zone and close to the valv\blueCNN{e}. These regions were identified as the main area of differences when comparing the local differences in terms of anatomical directions and strain patterns \cite{DiFolco:FIMH:2023}. Reaching a consensus on the RV anatomical directions is actually challenging \cite{Duchateau:JASE:2021}. While the \emph{long-axis} method we took as reference is simpler to compute, the \emph{heat diffusion} method was reported to provide more relevant outputs \cite{DiFolco:FIMH:2023}, in comparable computational times. 
\blueC{Also note that we focused \blueCNN{on} different ways to compute myocardial strain from an existing mesh sequence, previously obtained from segmentation and tracking by commercial software. Segmentation and tracking
methods, \blueCNN{the mesh creation process \cite{Kwan:JACCim:2024}, }and the imaging modality and its geometrical characteristics (e.g. 2D or 3D) may also cause additional differences in the quantification of the actual physical strain \cite{Muraru:JASE:2014}. Here, we preferred to remain in a more controlled and narrow setting (\blueCNN{d}ifferent definitions of strain applied to 3D echocardiographic data, complemented by experiments on a toy dataset of RGB images) to better focus on the \blueCNN{core method to apprehend definitional divergence}.}
\blueCNN{For RV echocardiography, there is a clear potential to build higher trust in 3D measurements, which are necessary given that the specific shape of the RV is insufficiently rendered in 2D. This also encourages the analysis of subtle local deformation patterns for disease characterization, beyond the quantification of peak global or wall-specific strain, as highlighted for the LV in 2D \cite{Duchateau:UMB:2020}.}

\blueCNN{Our methodology can be applied to other types of mesh data, including 3D RV meshes obtained from other cardiac imaging software (provided they can be exported and point-to-point correspondence can be obtained), and to other medical or imaging applications, as illustrated in the toy experiments in this paper. Application is possible as long as the inputs are of the same size, namely, point-to-point correspondence for meshes and uniform image size for images, and may require adapting the alignment method and metrics used to compare samples (see discussion of these specific aspects further in this section). Extension to pathological cases cannot be done at inference for the representation learning stage, but would instead require learning a new representation that includes such cases.} We hope that in the near future such \blueCNN{type of} computations could be a useful support to clinical interpretations, as already witnessed in other medical imaging fields \blueCNN{and a broader scope of uncertainties} \cite{Kompa:NPJDM:2021}.

We took advantage of \ac{PLS} to relate the strain \blueCNN{discrepancy maps} and the 3D \ac{RV} shape. We preferred this simple approach over non-linear ones, as local shape defects (e.g. folding) may be difficult to highlight on smooth shapes such as the \ac{RV} from 3D echocardiography \cite{Jia:ShapeMI:2018,Guigui:HandStats:2022}. Our analysis indicates that, beyond the anticipated volume and size variations when using a linear approach as \ac{PLS}, most shape differences and \blueCNN{discrepancy maps} align with anatomically challenging regions for the \ac{RV}. In particular, in contrast to the \ac{LV}, the asymmetrical and crescent shape of the \ac{RV} introduces distinct challenges. This may question the blind reliance on direction-dependent strain against non-directional one such as area strain, as previously reported \cite{Smith:JACC:2014}.

The manifold alignment method we used (\ac{MML}) suffers from two limitations: the metric to compare samples (here, the Euclidean distance between the data at all mesh points, considered as a column vector), and the reconstruction which is made a-posteriori. We preferred the simpler \ac{MML} approach to better focus on \blueCNN{modelling definitional divergence,} which constitutes the core of our methodological contribution. Nonetheless, our methodology \blueCNN{is} actually adaptable to any type of manifold alignment method, including those based on convolutional networks to better consider the local data structure (\blueCNN{which can be seen as using better metrics to compare samples ; our data would require using} graph convolutional networks \cite{Bronstein:SPM:2017}), and those with intrinsic reconstruction such as auto-encoders, which also require being adapted to shape data \cite{Ranjan:ECCV:2018}.
\blueCNN{Similarly, we preferred to keep the estimation of discrepancy maps from the latent samples simple (localized standard deviation across the reconstructed high-dimensional samples, meaning that each spatial location is considered independently). More advanced methods (e.g. reconstruction within a Bayesian framework) may provide finer estimation of discrepancy maps, in particular with better spatial consistency across the RV mesh, but at the price of higher computational cost and model complexity.}

Finally, manifold alignment falls within unsupervised representation learning and is therefore difficult to validate. We therefore designed experiments on toy synthetic and real data for which \blueCNN{the} locations and magnitude \blueCNN{related to definitional divergence} can be roughly guessed. Then, we progressively evaluated our methodology on real data with several configurations of increasing complexity (two and three close descriptors, then two correlated-but-different descriptors), thoroughly examining the \blueCNN{discrepancy maps} with the studied cardiac shapes and our prior knowledge on the data quality and disease.


\section{Conclusion}

We proposed a complete and original pipeline based on manifold alignment to \blueCNN{estimate the impact of definitional divergence, namely} different ways to compute a given descriptor, or complementary descriptors to represent a single patient's condition. Our approach \blueCNN{enables relevant assessment} locally, and is easily extendable to more elaborated types of manifold alignment based on graph neural networks and auto-encoders, and many other population analyses relying on heterogeneous high-dimensional descriptors.

\section*{Appendix A: Computational details of the manifold alignment}
\label{sec:AppendixMML}

We considered standard formulations for the matrices $\mathbf{W}^m$ and 
$\blueCN{\mathbf{V}^{mn}}$, 
namely:
\begin{equation}
    W_{ij}^m = \exp \frac{- \| \mathbf{x}_i^m - \mathbf{x}_j^m \|^2}{\sigma^2},
\end{equation}
where $\sigma$ is the width of the kernel defining the affinity matrix, and
\begin{equation}
    \blueCN{V_{ij}^{mn}} = \frac{\langle \mathbf{w_i^m} , \mathbf{w_j^n} \rangle}{ \| \mathbf{w_i^m} \| \| \mathbf{w_j^n} \| } \in [0,1],
\end{equation}
where $\mathbf{w_i^m}$ is the $i$-th row of the affinity matrix $\mathbf{W}^m$. Variants and discussion around such formulations \blueCNN{(including alternative measures of similarity between samples, in particular for the inter-modality correspondences)} can be found in \cite{Clough:PAMI:2019,DiFolco:MedIA:2022}.
\blueC{This formulation is somehow comparable to angular-based similarity matrices used for inter-modality correspondences, as proposed in multimodal clustering approaches \cite{Wang:MM:2014}.}

Equation \ref{eq:MML_large} can be reformulated using a block matrix $\mathbb{W} \in \mathbb{R}^{M K \times M K}$, whose diagonal and extra-diagonal blocks are the matrices $\mathbf{W}^m$ and $\blueCN{\mathbf{V}^{mn}}$, respectively \blueC{(see the demonstration in the Supplementary Material of \cite{DiFolco:MedIA:2022})}. This matrix formulation leads to solving:
\begin{equation}
\left.
    \begin{split}
    &E(\mathbf{Z}) = \text{tr}( \mathbf{Z}^T \mathbb{L} \mathbf{Z} ) \\
    \text{s.t.} \hspace{5mm} & \mathbf{Z}^T \mathbf{D}_{\mathbb{W}} \mathbf{Z} = \mathbf{I},
    \label{eq:MML_matrix}
    \end{split}
    \right\}
\end{equation}
where $\mathbb{L} = \mathbf{D}_{\mathbb{W}} - \mathbb{W}$ and $\mathbf{D}_{\mathbb{W}}$ is a diagonal matrix such that $\mathbf{D}_{\mathbb{W} ii} = \sum_j \mathbb{W}_{ij}$.
Eq.\ref{eq:MML_matrix} amounts at solving the generalized eigenvalue problem $\mathbb{L} \mathbf{f} = \lambda \mathbf{D}_{\mathbb{W}} \mathbf{f}$, where $\lambda$ and $\mathbf{f}$ stand for the eigenvalues and eigenvectors, respectively. In practice, we solve $\mathbb{P} \mathbf{f} = (1-\lambda) \mathbf{f}$, where $\mathbb{P} = \mathbf{D}_{\mathbb{W}}^{-1/2} \mathbb{W} \mathbf{D}_{\mathbb{W}}^{-1/2}$ is symmetric, which corresponds to working with the normalized graph Laplacian.

The coordinates $\mathbf{Z}$ correspond to the first eigenvectors associated to the first eigenvalues sorted by ascending order after removing the trivial case associated to the eigenvalue zero. As $\mathbb{W} \in \mathbb{R}^{MK \times MK}$, we obtain eigenvectors whose
rows $[K (m-1)+1 , K m]$ correspond to the low-dimensional coordinates for the $m$-th descriptor (the method therefore provides one latent space for each descriptor).

%

\section*{\blueC{Appendix B: modelling \blueCNN{definitional divergence} in the latent space}}
\label{sec:AppendixGaussian}

\begin{figure}[ht!]
	\centering
	\includegraphics[width=0.8\columnwidth]{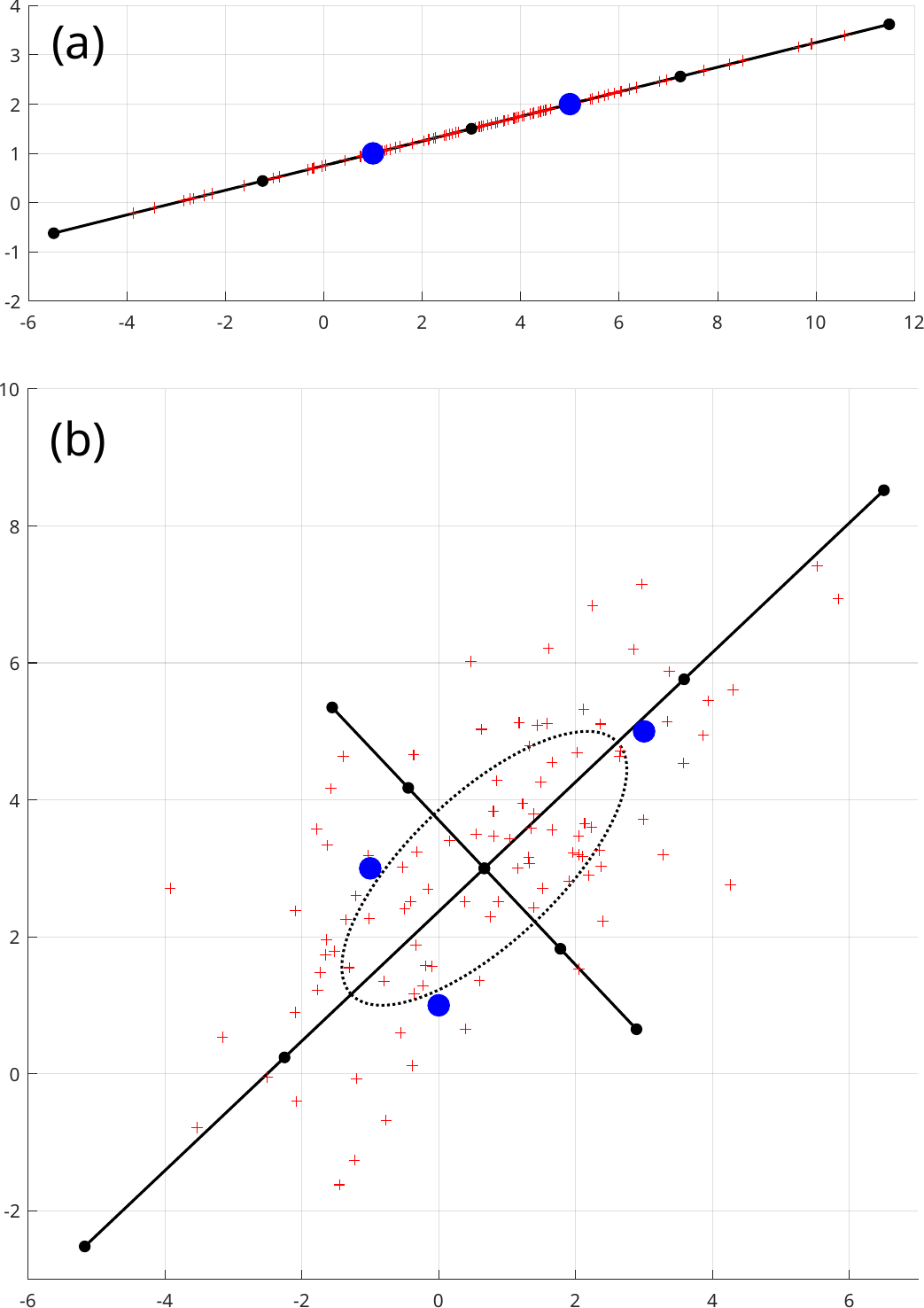}
	\caption{\blueC{Estimation of the principal axes (black lines / the black dots correspond to -3 to +3 standard deviations along these axes) of a Gaussian distribution from 2 (a) and 3 (b) samples (blue dots), and $N=100$ random samples generated from this distribution (red \blueCN{crosses}). In this illustrative experiment, the horizontal and vertical axes correspond to arbitrary dimensions. In our application, these could be seen as a projection of the low-dimensional samples into the hyperplane defined by the blue dots.}}
	\label{fig:AppB}
\end{figure}

\blueC{PCA estimates orthogonal axes corresponding to the principal directions of variance in the data.}
\blueC{Figure \ref{fig:AppB} illustrates its use to estimate a multivariate Gaussian defined by the set of points $\mathcal{Z}^i$, with $| \mathcal{Z}^i | = 2$ (Fig.\ref{fig:AppB}a) or $3$ (Fig.\ref{fig:AppB}b), and to sample $N=100$ new points from this distribution.}

\section*{Acknowledgments}
\footnotesize
The authors acknowledge the support from the French ANR (LABEX PRIMES of Univ. Lyon [ANR-11-LABX-0063] and the JCJC project ``MIC-MAC'' [ANR-19-CE45-0005]). Gabriel Bernardino was also partially supported by the FSE+ and “Agencia Estatal de investigacion” MICIU/AEI/ 10.13039/501100011033 (RYC2022-035960-I, PID2023-149959OA-I00). They are also grateful to P. Moceri (CHU Nice, France) for providing the imaging data related to the studied population, and to T. Dargent (LMD, IPSL Paris, France) for the initial computations on local coordinates and 3D \ac{RV} strain.
\normalsize

\bibliographystyle{IEEEtran}
\bibliography{ref}

\end{document}